\documentclass[conference]{IEEEtran}
\IEEEoverridecommandlockouts
\usepackage{cite}
\usepackage{amsmath,amssymb,amsfonts}
\usepackage{algorithmic}
\usepackage{graphicx}
\usepackage{textcomp}
\usepackage{xcolor}
\usepackage{tikz}
\usepackage{lipsum}
\usepackage{booktabs}
\usepackage{xcolor}
\usepackage{multirow}
\def\BibTeX{{\rm B\kern-.05em{\sc i\kern-.025em b}\kern-.08em
    T\kern-.1667em\lower.7ex\hbox{E}\kern-.125emX}}
\begin{document}

\newcommand{\mymethod}{{\sc CrypText}}

\title{{\mymethod}: Database and Interactive Toolkit of Human-Written Text Perturbations in the Wild}

\newcommand\copyrighttext{%
  \footnotesize \textcolor{red}{\textbf{DISCLAIMER! THIS PAPER CONTAINS EXAMPLE TEXTS THAT ARE OFFENSIVE IN NATURE}}}
\newcommand\copyrightnotice{%
\begin{tikzpicture}[remember picture,overlay]
\node[anchor=south,yshift=18pt] at (current page.south) {{\copyrighttext}};
\end{tikzpicture}%
}


\author{Thai Le \hspace{0.3in} Ye Yiran \hspace{0.3in} Yifan Hu{*} \hspace{0.3in} Dongwon Lee\vspace{0.05in} \\
\hspace{-0.46in} The Pennsylvania State University \hspace{0.65in} Yahoo Inc.{*}\\
\{tql3,yby5204,dongwon\}@psu.edu \hspace{0.2in} yifanhu@yahooinc.com{*}}

\maketitle

\begin{abstract}
User-generated textual contents on the Internet are often noisy, erroneous, and not in correct grammar. In fact, some online users choose to express their opinions online through carefully {\em perturbed} texts, especially in controversial topics (e.g., politics, vaccine mandate) or abusive contexts (e.g., cyberbullying, hate-speech). However, to the best of our knowledge, there is no framework that explores  these online ``human-written" perturbations (as opposed to algorithm-generated perturbations). Therefore, we introduce an interactive system called {\mymethod}. {\mymethod} is a data-intensive application that provides the users with a database and several tools to extract and interact with human-written perturbations. Specifically, {\mymethod} helps look up, perturb, and normalize (i.e., de-perturb) texts. {\mymethod} also provides an interactive interface to monitor and analyze text perturbations online. A short demo video is available at: \textit{https://youtu.be/8WT3G8xjIoI}.
\end{abstract}

\begin{IEEEkeywords}
perturbations, database, machine learning
\end{IEEEkeywords}

\section{Introduction}

On popular social platforms such as Twitter, Reddit or Facebook, users sometimes slightly change the spelling of a text, without changing its original semantic meaning, for various reasons.
For instance, the following sentences contain real perturbed texts (underlined):
\textit{``The \underline{democRATs} responsible for their attempted race war,"} \textit{``A fake tree burned and \underline{RepubLIEcans} are calling for,"} or \textit{``Thinking about \underline{suic1de}."} 
These perturbations are often users' deliberate attempts to emphasize their opinions implicitly or explicitly (e.g., democRATs and RepubLIEcans), or 
to censure offensive or sensitive wording to avoid platforms' censorship (e.g., suic1de).
Moreover, we also found that religion or nationality related perturbations  (e.g., ``muslim"$\rightarrow$``mus-lim" and ``chinese"$\rightarrow$``chi-nese") often appear in the context of   cyberbullying, racism, or hatespeech on platforms.

These human-written perturbations are a useful resource for both researchers and practitioners as they provide inductively-derived (i.e., observable) attacks by humans, as opposed to deductively-derived (i.e., hypothesized) adversarial attacks by machine learning (ML) algorithms.
Although such human-written perturbations occur frequently online, however, there has been no tool that helps discover, study, and analyze these human-written perturbations. Therefore, in this demo work, we propose {\mymethod}, a system that detects and extracts human-written perturbations from social platforms in a large scale. 
The contributions of {\mymethod} can be summarized as follows:
\begin{itemize}
    \setlength\itemsep{-2pt}
    \item {\mymethod} is uniquely equipped with a dictionary of over 2M human-written tokens that are categorized into over 400K unique phonetic sounds. {\mymethod} is constantly learning new perturbations from social platforms of
    Reddit and Twitter, providing users with up-to-date perturbations. {\mymethod} also enables the users to discover perturbations from input texts and normalize or de-perturb them (i.e., correct perturbed texts). 
    \item {\mymethod} also facilitates the evaluation of the robustness of textual ML models on noisy user-generated contents by perturbing inputs with realistic human-written (and not machine-generated) perturbations. {\mymethod} also helps analyze the usage patterns of human-written perturbations and their associated sentiments online.
    \item {\mymethod} is also beneficial for a variety of users and audiences including NLP researchers, linguists and social scientists. {\mymethod} will be open-sourced.
\end{itemize}

\section{Literature Review}

\subsection{Definition of Text Perturbations}

Previous research on the use of text perturbations, are only limited within the context of \textit{machine-generated} adversarial texts in the literature. Particularly, they propose different  algorithms to \textit{minimally} manipulate an input text $x$ to generate an adversarial text $x'$ that can fool a target ML model $f(\cdot)$ such that $f(x'){\neq}f(x)$ and $x'$ still preserves the semantic meaning of $x$ as much as possible.
Then, we can define \textit{text perturbations} of $x$ (which are different from adversarial texts), regardless of whether they are generated by machine or written by human, as a collection of manipulated texts $P_x{=}\{x'_1, x'_2,..x'_N\}$ yet \textit{\textbf{without}} the adversarial condition (i.e., $f(x')\neq f(x)$).
From now on, we will use the notation $x'\in P_x$ to denote a perturbation (and \textit{not} an adversarial example) of $x$. Moreover, we specifically focus on character-level perturbations since they are intuitive to human.
%


\subsection{Machine-Generated  Perturbations}
There are several automatic character-level text manipulation algorithms proposed in the adversarial NLP literature. These manipulation strategies include swapping, deleting a character in a word (e.g., ``democrats"$\rightarrow$``demorcats", TextBugger~\cite{li2018textbugger}), replacing a character by its most probable misspell (e.g., ``republicans"$\rightarrow$``rwpublicans", TextBugger~\cite{li2018textbugger}), replacing a character by another visually similar digit or symbol (e.g., ``democrats"$\rightarrow$``dem0cr@ts" TextBugger~\cite{li2018textbugger}), an \textit{accent} (e.g., ``democrats"$\rightarrow$``d\.emocr\=ats", VIPER~\cite{VIPER}), or a \textit{homoglyph} (DeepWordBug~\cite{gao2018black}). Although these strategies are shown to be very effective in flipping textual ML models, they are hypothesized based on some known vulnerabilities of a target textual model $f(\cdot)$, and \textit{not inductively} derived from actual real-life observations. Thus, their applications are limited only to adversarial NLP settings.

\subsection{Human-Written Perturbations}\label{sec:human_written_text}
From analyzing several public text corpus, we observe that machine-generated and human-written perturbations are very different. For example, human often add another layer of meaning on words by emphasizing a part of it (e.g., ``democRATs", ``repubLIEcans"). Moreover, hyphenation are also used to perturb a word (e.g., ``mus-lim", ``vac-cine"). Different from machine, humans perturb words by using emoticons, repeating characters (e.g., ``porn"$\rightarrow$``porrrrn"), and using phonetically similar characters (e.g., ``depression"$\rightarrow$``depresxion"). Human's manipulation strategies can also be very creative and require some implicit context to understand (e.g., ``democRATS"). Often, human-written perturbations also do not manifest any fixed rules. Hence, it is very challenging for an algorithm to systematically generate all human-written perturbations. As these strategies become more prevalent online, conventional ML models that often assume clean English input texts will face difficulties with noisy texts in real-life, especially models that deal with controversial topics (e.g., politics, vaccine mandate), or abusive contents (e.g., cyberbullying, hate-speech).


\section{Demonstration of {\mymethod} System}
As the first step toward understanding human-written perturbations online, we design and propose the {\mymethod} that discovers, utilizes and monitors human-written perturbations online. {\mymethod} has four main functions: (1) looking up perturbations, (2) normalizing perturbations, (3) manipulating texts using human-written perturbations, and (4) visualizing and analyzing online contents through the lens of text perturbations. All functions are provided via a graphical-user-interface (GUI) on top of a database (DB) system
and a set of public APIs. We elaborate each functions and its use cases below. 

\begin{figure}[tb!]
    \centering
    \includegraphics[width=0.4\textwidth]{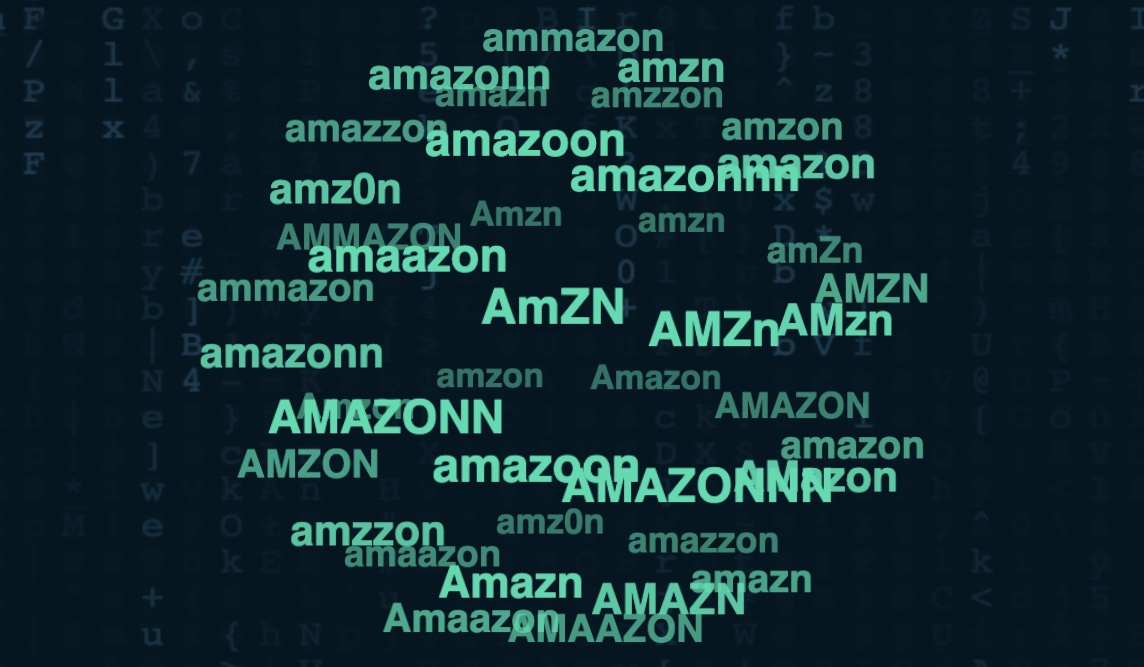}
    \caption{\textit{Look Up} function displays human-written perturbations for the input token ``amazon" in a form of an interactive 3D spherical word-cloud.}
    \label{fig:lookup}
    \vspace{-10pt}
\end{figure}

\subsection{Human-Written Token Database}
{\mymethod} relies on the database of human-written tokens of both correctly-spelled English words and their potential perturbations. This database stores all of the raw case-sensitive tokens found from several public datasets. Most of them are originally curated for detecting abusive materials online (e.g., rumors~\cite{kochkina2018all}, hatespeech~\cite{gomez2020exploring}, cyberbullying~\cite{wulczyn_thain_dixon_2017}) and often contain many perturbations. To curate the database, we first tokenize each sentence in the corpus, and encode their sounds using a customized version of the \textsc{Soundex} algorithm~\cite{stephenson1980methodology}. 

Given a token, \textsc{Soundex} first fixes the first character, then matches the remaining characters one-by-one following a pre-defined rule set (e.g., \{b, f, p, v\}{$\rightarrow$}``1"). However, the original \textsc{Soundex} algorithm cannot recognize some visually-similar character manipulations that are frequently found in human-written perturbation texts (e.g., ``l"$\rightarrow$``1", ``a"$\rightarrow$``@", ``S"$\rightarrow$``5"). Moreover, fixing the first character can result in many mismatches (e.g., both ``losbian" and ``lesbian" share the same code ``L215" yet has distinct sounds). To overcome these, we thus customize the \textsc{Soundex} algorithm to encode visually-similar characters the same. We also introduce the \textit{phonetic level} parameter $\mathbf{k}$ and fix the first $\mathbf{k}+1 (\mathbf{k}{\geq}0)$ characters as part of the resulted encodings (e.g., ``losbian"$\rightarrow$``LO245", ``lesbian"$\rightarrow$``LE245" with $\mathbf{k}{\leftarrow}$1). All the extracted data using the modified \textsc{Soundex} algorithm are then stored as several hash-map $\mathcal{H}_\mathbf{k}$ with different values of $\mathbf{k}{\leq}2$ in the database. Each key and value of $\mathcal{H}_\mathbf{k}$ is a unique \textsc{Soundex} encoding and a set of all tokens sharing the same encoding at the phonetic level $\mathbf{k}$. Table \ref{tab:example} shows an example of $\mathcal{H}_1$ extracted from a set of three sentences.

\renewcommand{\tabcolsep}{7pt}
\begin{table}[tb]
    \centering
    \footnotesize
    \begin{tabular}{lr}
        \toprule
        \textbf{Key} & \textbf{Value} \\
        \cmidrule(lr){1-2}
        TH000 & \{the, thee\} \\
        DI630 & \{dirty, dirrrty\} \\
        RE4425 & \{republicans, repubLIEcans, republic@@ns\} \\
        \bottomrule
    \end{tabular}
    \caption{Example of extracted hash-map $\mathcal{H}_1$ from a corpus of two sentences \textit{``the dirrty republicans"}, \textit{``thee dirty repubLIEcans"}, \textit{``the dirty republic@@ns"}}
    \label{tab:example}
    \vspace{-15pt}
\end{table}


\subsection{\textit{Look Up}: Discovering Text Perturbations}\label{sec:lookup}

\noindent \textbf{Functionality.} {\mymethod} allows users to query to the DB a set of perturbations $P_x$ for a specific token $x$ in the form of an interactive 3D \textit{word-cloud} (Figure. \ref{fig:lookup}). To do this, {\mymethod} relies on what we called the \textit{SMS} property to characterize text perturbations. Specifically, we define a perturbation as a token that has phonetically similar \underline{S}ound, is perceived with the same \underline{M}eaning as its original word, yet
has different \underline{S}pelling.
For example, we consider ``demokRATs" as a perturbation of ``democrats" because they spell differently yet sound the same, and human readers can perceive both with the same meaning, 
especially when placed in similar contexts (e.g., \textit{``Biden belongs to the democrats"} and \textit{``Biden belongs to the demokRATs"}). However, there are no automatic mechanisms that can measure the semantic similarity between two words of different spellings, especially when misspelled tokens are often encoded as a out-of-vocabulary (OOV). To overcome this, we utilize their Levenshtein edit-distance $\mathbf{d}$ in addition to their similarity in \textsc{Soundex} encodings as a proxy for semantic similarity. Intuitively, two tokens share the same semantic meaning if they phonetically sound the same and are separated by a sufficiently small Levenshtein distance. 
Given the example in Table \ref{tab:example}, a search using the query $x{\leftarrow}\mathrm{``republicans"}$ with $\mathbf{k}{=}1, \mathbf{d}{=}1$ results in $P_x\leftarrow$\{republicans, repubLIEcans\}. Figure \ref{fig:lookup} shows an example output of the \textit{Look Up} function. We manually set $\mathbf{k}{=}1, \mathbf{d}{=}3$ by default. Advanced users will be able to adjust these parameters through a provided API.

\noindent \textbf{Use Cases.} One notable use case of \textit{Look Up} is \textit{keyword enrichment}, which provides additional queries to search for sensitive topics online. Specifically, it enables  users to search for contents that otherwise could have been censured or unreachable with correctly-spelled keywords. For example, only 67\% of the tweets found by Twitter's search API from Nov. 2021 using keyword ``democrats" has negative sentiment, while that number is much higher of 87\% if a search query also includes the perturbations of ``democrats" from the \textit{Look Up} function. We also observe the same results for ``republicans" (66\% v.s. 84\%) and ``vaccine" (46\% v.s. 61\%). We also found that most tweets that contain the perturbations of ``vaccine" focus more on either pushing back vaccine mandate or raising concerns on its safety and effectiveness, compared with the ones that are not perturbed.

\subsection{\textit{Normalization}: Detecting and De-Perturbing Texts} 
\begin{figure}[tb!]
    \centering
    \includegraphics[width=0.4\textwidth]{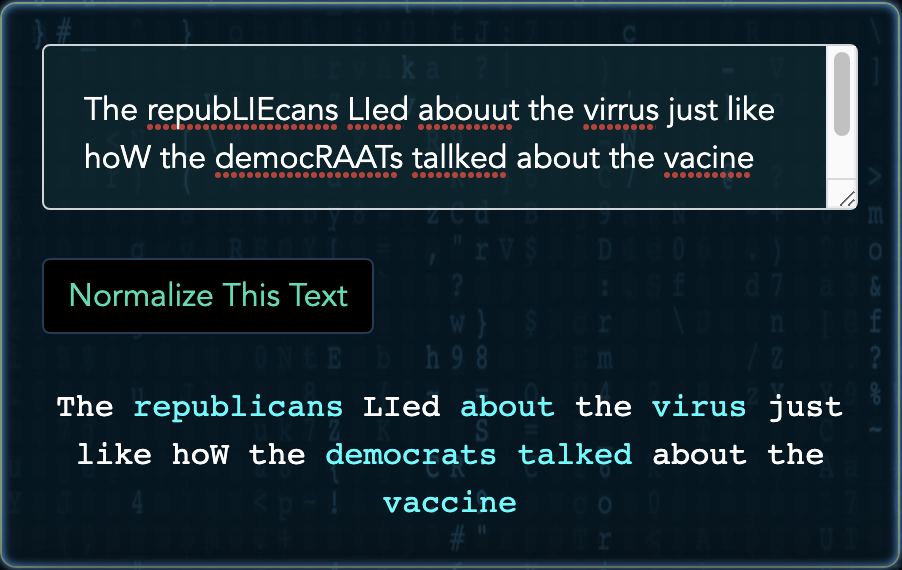}
    \caption{\textit{Normalization} function displays the normalized version of the input with corrected tokens being highlighted. The GUI also shows a popup describing a token before and after normalization.}
    \label{fig:normalization}
\end{figure}

\noindent \textbf{Functionality}. This function detects and corrects text perturbations from an input $x$ by leveraging the SMS property described in the \textit{Look Up} function. Specifically, for each $i$-th token $x_i$ in $x$, {\mymethod} retrieves a set of potential matching English words for $x_i$ given hyper-parameter $\mathbf{k}, \mathbf{d}$. A token $w^*$ is a valid perturbation if there exists an English word that shares the same sound encoding as $x_i$ at phonetic level $\mathbf{k}$ (i.e., belong to the same hash-map $\mathcal{H}_\mathbf{k}$), within an upper-bound edit-distance $\mathbf{d}$. However, there can be several candidate English words that associate with the token $x_i$. In this case, {\mymethod} further ranks them by approximating how they fit into their surrounding local context of $x_i$ in $x$. Specifically, we utilize a large pre-trained masked language model $G$ to calculate a \textit{coherency} score. Intuitively, this score is calculated by utilizing the pre-trained power of a complex language model such as BERT~\cite{devlin2018bert} to calculate how likely $w^*$ appears in the immediate context of $x_i$. Then, {\mymethod} outputs the most probable {\em de-perturbed} candidate for each $x_i{\in}x$ on its GUI (Figure. \ref{fig:normalization}). Advanced users can also retrieve all candidates $w^*$ and their coherency scores via a provided API.

\noindent \textbf{Use Cases.} There are two notable uses cases, namely (1) de-noising inputs of textual ML models and (2) usage as an additional predictive signal for ML pipelines. Given that the majority of ML models are often trained only on clean English corpus, {\mymethod} can be used to correct all possible human-written perturbations in the training corpus.
Moreover, the presence of perturbations within a sentence can also inform potential adversarial behaviors from its writer, especially those  offensive or controversial perturbations (e.g., cyberbullying, political discourse, vaccine mandates, spreading disinformation), as part of a ML pipeline.

\begin{figure}[tb!]
    \centering
    \includegraphics[width=0.4\textwidth]{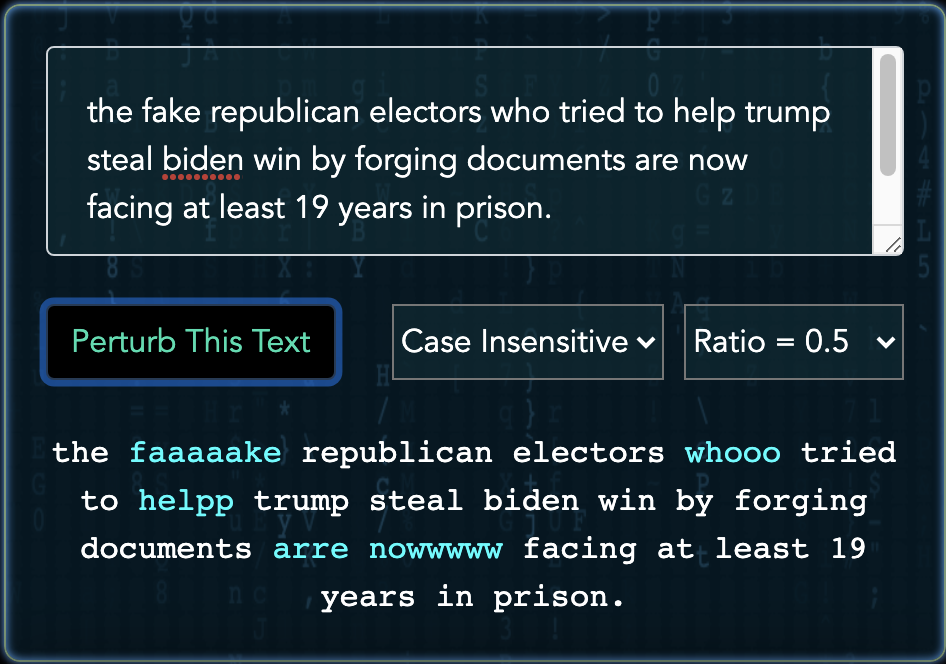}
    \caption{{\mymethod} perturbs an tweet with highlighted perturbations.}
    \label{fig:perturbation}
    \vspace{-10pt}
\end{figure}

\subsection{{\textit{Perturbation}: Text Manipulation using Human-Written Perturbations}} 

\noindent \textbf{Functionality}. This function helps manipulate an input text using either case-sensitive or case-insensitive human-written perturbations at a user-specified manipulation ratio $\mathbf{r}$ (e.g., 15\%, 25\%, 50\%). To do this, given an input text $x$, {\mymethod} first randomly samples a subset of tokens in $x$ according to the selected $r$ value. Then, it iteratively replaces each of the selected tokens by a perturbation that is randomly selected from the \textit{Look Up} function's output (Figure. \ref{fig:perturbation}). 

\begin{figure}[tb]
    \centering
    \includegraphics[width=0.4\textwidth]{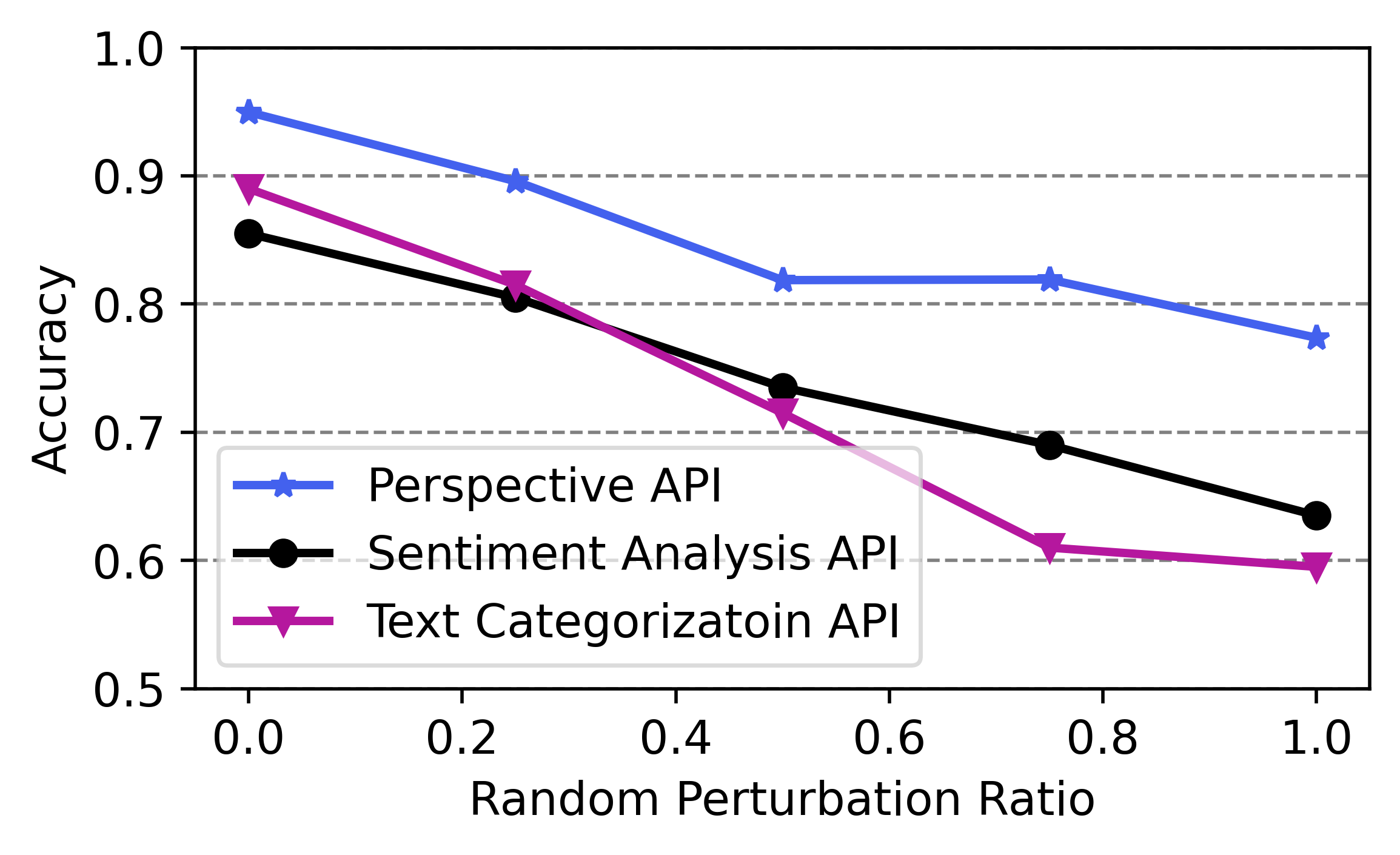}
    \caption{Accuracy of Google's NLP APIs on texts perturbed by {\mymethod}.}
    \label{fig:google}
    \vspace{-10pt}
\end{figure}

\noindent \textbf{Use Case.} Compared with other text perturbation methods in the literature (e.g., TextBugger~\cite{li2018textbugger}, VIPER~\cite{VIPER}, DeepWordBug~\cite{gao2018black}), perturbations utilized by {\mymethod} are guaranteed to be observable in human-written texts. This makes {\mymethod} a more realistic tool to evaluate the robustness of textual ML models on noisy user-generated texts. This implication is applicable not only to NLP classification models, especially those that are built for abusive content moderation, but also on models developed for other tasks such as natural language inference and Q\&A. In fact, many of the industrial APIs are not well-equipped to deal with noisy human-written texts. Figure \ref{fig:google} shows that the \textit{Perspective} toxic detection, sentiment analysis, and text categorization API from Google Cloud all suffer from texts perturbed by {\mymethod}. Especially, the popular \textit{Perspective} API's accuracy drops nearly 10\% when only 25\% of input texts are perturbed. {\mymethod} also dedicates an \textit{ML benchmark} page that frequently updates our evaluation of publicly available NLP APIs and models on noisy human-written texts.

\subsection{\textit{Social Listening}: Monitoring Human-Written Perturbations Online}
\noindent \textbf{Functionality.} This function enables users to monitor the use of human-written perturbations online, especially on social platforms such as Reddit (via \textit{PushShift API\footnote{https://github.com/pushshift/api}}). Specifically, given a list of English words, {\mymethod} first searches on the social platforms all the contents using their perturbations as queries. Then, it aggregates and displays the usage patterns of each individual perturbation in both frequency and sentiment through interactive timeline charts. We refer the readers to the attached video for the full visualization of this function.
\noindent \textbf{Use Case.} As described in the use case of \textit{Look Up} function (Sec. \ref{sec:lookup}), using perturbations as queries help capture a more comprehensive view on a specific topic, especially one that is sensitive or divided. \textit{Social Listening} helps achieve this by capturing contents that are often contrast or opposite to conventional opinions, which otherwise could be censured or unreachable on non-perturbed texts. Likewise, gatekeepers of social platforms also can utilize this function for better content moderation, especially in detecting and removing abusive texts on web (e.g., cyberbullying, racism, calling for violence and terrorism),  many of which are often intentionally written with misspellings to evade automatic detection.

\subsection{An Architectural Design}
\begin{figure}[t!b]
    \centering
    \includegraphics[width=0.35\textwidth]{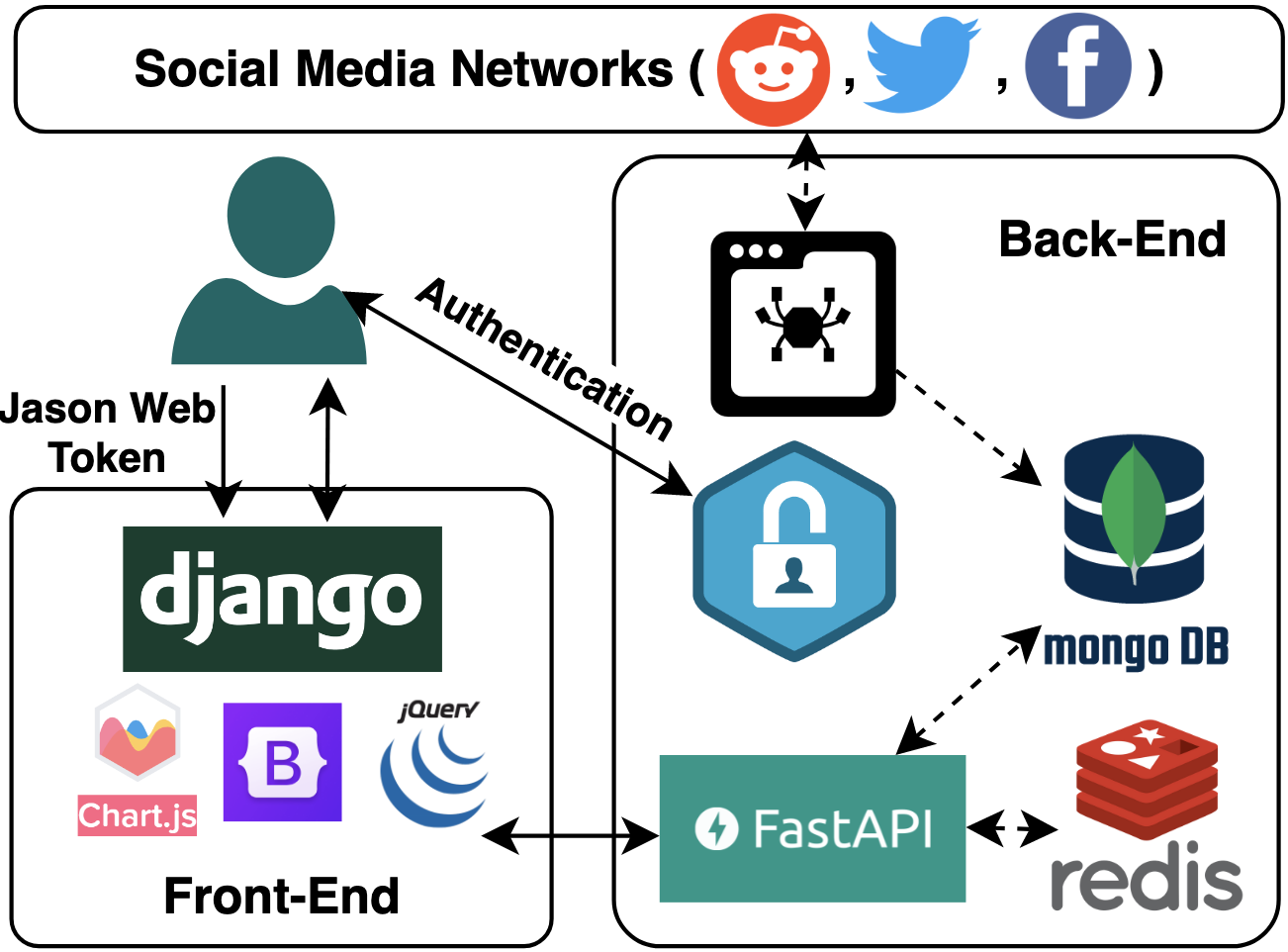}
    \caption{An implementation of \mymethod.}
    \label{fig:architecture}
    \vspace{-10pt}
\end{figure}

Figure \ref{fig:architecture} illustrates the architecture of the {\mymethod} implementation. {\mymethod} interacts with users via a GUI (i.e, website) and several function APIs. The back-end of {\mymethod} is programmed with the \textit{Django} and \textit{FastAPI} framework in \textit{python} language. Its front-end utilizes several powerful visualization tools such as \textit{chart.js}, \textit{TagCloud.js}, and \textit{dataTables.js} to provide the end-users with an interactive experience. All functions of {\mymethod} are equipped with secured public APIs, allowing users to utilize \textit{Look Up}, \textit{Normalization} and \textit{Perturbation} in bulks. Accessing such APIs requires an authorization token that will be provided upon request. All of the data utilized by {\mymethod} are stored in a MongoDB database. Since some queries might take a longer time to process, a \textit{Redis cache} is adapted to temporarily store and re-use recent queried results to enhance the user experience. Furthermore, we set up a crawler that regularly collects recent tweets (via Twitter's public stream API\footnote{\url{https://developer.twitter.com/}}) to continually enrich {\mymethod}'s database with novel perturbed tokens online.

\section{Limitation and Conclusion}
Currently, the \textit{Social Listening} function is limited to Reddit data and we plan to support other platforms in future. To the best of our knowledge, {\mymethod} is the first interactive database system that helps discover and utilize human-written text perturbations online. Database provided by {\mymethod} can also help evaluate the robustness of textual ML models on noisy human-written inputs and uncover sensitive online discussions. Especially, {\mymethod} constantly updating its DB with new perturbations via Twitter's stream API. 



\bibliographystyle{IEEEtran}
\bibliography{main}

\end{document}